\documentclass[11pt]{article}
\usepackage{geometry}
\usepackage{coling2020}
\usepackage{times}
\usepackage{url}
\usepackage{latexsym}
\usepackage{microtype}
\usepackage{appendix}

\colingfinalcopy 

\title{Duluth at SemEval-2020 Task 12:\\Offensive Tweet Identification in English with Logistic Regression}

\author{Ted Pedersen \\
Department of Computer Science \\
University of Minnesota\\
Duluth, MN 55812 USA \\
{\tt tpederse@d.umn.edu}}

\date{}

\begin{document}
\maketitle
\begin{abstract}
This paper describes the Duluth systems that participated in SemEval--2020 Task 12,  Multilingual Offensive Language Identification in Social Media (OffensEval--2020). We participated in the three English language tasks. Our systems provide a simple Machine Learning baseline
using logistic regression. We trained our models on the distantly supervised training data made available by the task organizers and used no other resources. As might be expected we did not rank highly in the comparative evaluation: 79\textsuperscript{th} of 85 in Task A, 34\textsuperscript{th} of 43  in Task B, and 24\textsuperscript{th} of 39 in Task C. We carried out a qualitative analysis of our results and found that the class labels in the gold standard data are somewhat noisy. We hypothesize that the extremely high accuracy ($>$ 90\%) of the top ranked systems may reflect methods that learn the training data very well but may not generalize to the task of identifying offensive language in English. This analysis includes examples of tweets that despite being mildly redacted are still offensive.
\end{abstract}

\section{Introduction}

\blfootnote{
    %
    %
    \hspace{-0.65cm}  
     This work is licensed under a Creative Commons Attribution 4.0 International License. License details: \url{http://creativecommons.org/licenses/by/4.0/}.
}

The goal of the OffensEval--2020 task \cite{Offenseval2020} is to identify offensive language in tweets, and to determine if specific individuals or groups are being targeted. We relied on traditional Machine Learning methods implemented in Scikit \cite{Scikit-learn} to build logistic regression classifiers from distantly supervised training examples of offensive tweets \cite{Rosenthal2020}. Our methods are well known and so will only be described briefly. Instead, our primary focus in this paper is on a post-evaluation qualitative analysis of both our results and the underlying task data.

Identifying offensive, abusive, and hateful language is a challenging problem that is drawing increasing attention both among the general public and in the research community (e.g.,\cite{FortunaN18,SchmidtW17}). These are difficult problems since what is offensive depends not only on the words being used but also on the situation in which they occur. Whether something is offensive may depend on answers to questions like : Is the source of such language in a position of power? Is the target a member of a marginalized group? Is there a difference in age, race, religion, or social status of the source and the target? The number of real-world factors that may determine if language is offensive is impossible to enumerate, and leads to a high degree of polysemy in candidate offensive words that can only be unraveled by considering the social situation in which they are used. For example, words that are usually considered slurs may be used within a marginalized group as a means of bonding or identifying. The same is true of profanity, which when used within an in-group or among friends may not be offensive, while the same language directed at an outsider might well be offensive \cite{WaseemTB2018,Sapetal2019,Wiegandetal2019}. 

\section{Experimental Data}
\label{section:data}

OffensEval--2020 is made up of three tasks that went through the final evaluation stage in late February and early March 2020. Task A is to classify a tweet as offensive (OFF) or not (NOT). Task B takes the tweets identified as OFF from Task A and determines if hey are targeted insults (TIN) or not (UNT). Task C considers the targeted insult tweets from Task B and classifies them as being directed against an individual  (IND), group (GRP) or other entity (OTH). There is a cascading relationship between the tasks, where Task C requires the output of Task B, and Task B requires the output of Task A. This is the same set of labels and tasks as used in OffensEval--2019. 

OffensEval--2020 provided a large corpus of training tweets known as SOLID that was created via distant supervision \cite{Rosenthal2020}. Task A included 9,089,139 tweets, Task B provided 188,974 and Task C 188,973. The distant supervision used to create SOLID was seeded with the the manually labeled OLID corpus from OffensEval--2019  \cite{zampieri2019predicting}. OLID is a much smaller corpus of 13,240 training tweets and 860 test tweets which was also available to OffensEval--2020 participants. 

SOLID and OLID are different in that tweets in OLID are labeled with categories whereas in SOLID tweets are scored on a continuous scale of 0 to 1 to reflect the collective judgment of the models used as a part of distant supervision. As such the SOLID data did not provide a specific indication as to the boundaries between categories. 

\section{Assigning Labels to the Experimental Data} 
\label{section:labels}

We made a few significant decisions early on regarding the data. First, we elected not to use the 2019 OLID data. We participated in OffensEval--2019 and observed some potential inconsistencies in the OLID training data \cite{Pedersen19}. We also felt that since SOLID was seeded with OLID that there would be no particular advantage to also using OLID. There is a tradeoff here between using a smaller amount of manually annotated data (OLID) versus a much larger sample of potentially noisier distantly supervised data (SOLID). Second, we decided to directly map the 2020 SOLID training data to categories. This required us to draw somewhat perilous and arbitrary boundaries through a real valued space for each task. 

We studied the distribution of scores in Task A and observed that the median in the SOLID training data was .25, and that the standard deviation was .185. This suggested that the vast majority of tweets were not considered offensive, and so we selected the value of .8 for our cutoff. Based on our manual review of the training data we felt that tweets in Task A with a score greater than .8 were very likely to be offensive. We realized that we could set this boundary lower (perhaps .75 or .70) and still include many offensive tweets, but wanted to choose a boundary that might at least give the possibility of high precision results. 

After making this cutoff, only about 4\% (356,811) of the training tweets were labeled as offensive. We used this same threshold with Task B and C. For Task B, any tweet with a score less than .2 was considered targeted.  In Task C separate scores were provided for each of the three possible categories, so any tweet with a score greater than .8 was considered to be targeted against an individual, group or other.

\section{Methods}

The Duluth system is a slightly modified version of a tweet emoticon classification system developed for SemEval--2018 \cite{JinP18}. This system does some light pre-processing that largely leaves the tweets intact (but does basic cleaning and regularization of punctuation) and then identifies unigrams and bigrams as candidate features. We took the same approach for each task, where we learnt a logistic regression classifier from the training data where category boundaries were set as described in Section \ref{section:data}. 

\section{Experimental Results}

The official rankings in OffensEval were based on macro--averaged F1, and we also report accuracy and per-class precision, recall, and $F1$ scores.

The results of the Duluth systems were significantly below top ranked systems. For Task A our macro F-1 score was .7713 while the top ranked system attained .92226, for Task B we obtained .53816 versus a top rank of .74618, and for Task C the Duluth system scored .57437 against .71450.

That said, our qualitative analysis of our results suggests that there is some noise in the class labels of the gold data, and so we are uncertain if significantly higher accuracy would translate into meaningful performance on the task of identifying offensive language in English. In the confusion matrices that follow, the distribution of gold answers (ground truth) are shown on the columns, and the system predictions are on the rows. 

\subsection{Task A}

Task A asks whether a tweet is offensive (OFF) or not (NOT). 
This task had the largest amount of training data (9,0899,140 examples), of which we considered 4\% offensive (OFF) and 96\% not (NOT). 

We found that the Duluth system was very conservative in predicting a tweet as OFFensive, assigning that label to only 16\% (603) of the 3,887 test instances. This resulted in  a classifier that achieved high precision in identifying offensive tweets. We hypothesized that this would be the case when we set the cutoff of .8 for offensive tweets (see Section \ref{section:labels}). 

Table \ref{table:taskaresults} shows that the most significant confusion for our system was disagreement with the gold standard over what was offensive. Our system considered fully half (540) of the gold standard offensive tweets to be NOT offensive (shown in bold face in Table \ref{table:taskaresults}). 

\begin{table}[h]
\centering
\begin{tabular}{lrrr|rrr}
     &   OFF &   NOT & & P & R & F1 \\ \hline
 OFF &   540 & 63 & 603 & .896 & .500 & .642 \\
 NOT &  {\bf 540}  & 2,744 & 3,284 & .164 & .978 & .281 \\ \hline
     &  1,080  & 2,807 & 3,887 &  &  &  \\ \\
\end{tabular}
\caption{Task A Duluth : Logistic Regression, accuracy = .845, majority classifier = .722}
\label{table:taskaresults}
\end{table}

\begin{table}[h]
\centering
\begin{tabular}{l|l}
A1601 & @USER Not ashamed to say it, but i jacked off to the thought of having \\
& sex with this guy.
What i wouldnt do to go to bed with him \\
A848 & @USER @USER 100\% liar Pakistan is the country which damaged the image of Muslim \\
& nation around the world \\
A388 & dear white men, pink dick is out of season \\
A2403 & @USER Back to Somila. You hate America. You hate her people. \\
& You promote terror. We don't need you. \\
A1267 & It’s always that 1 b**ch that wanna be related to everybody\\
A1448 OFF? & @USER 6’9 with the soul of a 7’4 n**ga \\
A2127 OFF?? & That. Ain’t. Got. Shit. To. Do. With. Me. Idc who came before me, 
it’s IRRELEVANT. \\
& cause I’ve been b**ched about it before   Do you feel it’s a must to disclose who’ve you slept \\
& with to someone during the talking stage? \\
A2816 OFF?? & @USER He will fight them on the beaches ... what is it with these
pathetic little boys \\
& and their dreams of glory? \\
A3005 OFF?? & @USER @USER Such left leaning bullshit rules that don't apply to 
the LEFT as they \\
& punish the RIGHT instead \\
A237 OFF?? & @USER Spoiler alert.. George was set up by the corrupt Dems and Hillary, \\
& just like President Trump! \#FisaBringsDownTheHouse \\
\end{tabular}
\caption{Task A OFF-NOT category examples - gold standard indicates offensive while Duluth system predicts not offensive.}
\label{table:offnot}
\end{table}

Given the significant disagreement between gold OFFensive and predicted NOT, we carried out a qualitative analysis of randomly selected test instances that fell into this category. Specifically, we randomly selected 10 instances from the OFF-NOT category 10 different times. 
Table \ref{table:offnot} shows one of these ten trials (which was itself randomly selected). 
We've indicated (with OFF??) five instances where we believe a case can be made that the tweet is not offensive. Overall in our ten trials we found anywhere from two to five instances per trial where there could be reasonable doubt as to whether the tweet was offensive. We carried out a similar analysis with the three other cross-classification categories for Task A and show examples of that in Tables~\ref{table:offoff}, \ref{table:notnot}, and \ref{table:notoff} in the Appendix.

We noted in-group uses of the N-word and the B-word seemed to be automatically considered offensive (in all tasks). A1448 is an example, where the tweet may in fact be intended as a compliment. The automatic classification of profanity and slurs as offensive is a known and significant problem, since this can lead to all in-group speech among members of a marginalized group as being unfairly labeled as offensive \cite{WaseemTB2018,Sapetal2019,Wiegandetal2019}. 

These observations suggest that a highly accurate classifier trained on this data may simply be learning anomalies of this sort and may not generalize well to the problem of identifying offensive language.

\subsection{Task B}

Task B takes the tweets labeled as OFFensive in Task A and determines if they are targeted insults (TIN) or not (UNT). We can see in Table \ref{table:taskbresults} that the Duluth system was again very conservative, this time in considering a tweet to be targeted. In this task our precision has fallen fairly dramatically, and our accuracy lags behind even a majority classifier.  Our system disagrees with the gold standard 70\% of the time (in 593 of 850 tweets, shown in bold face in Table \ref{table:taskbresults}) and is far less likely to consider a tweet a targeted insult. 

\begin{table}[h]
\centering
\begin{tabular}{lrrr|rrr}
     &   UNT &   TIN & & P & R & F1 \\ \hline
 UNT &  533  & {\bf 593}     & 1,126 & .473 & .932 & .628 \\
 TIN &  39   & 257     & 296 & .132 & .068 &  .090\\ \hline
     & 572   & 850    & 1,422 &  &  &  \\
\end{tabular}
\caption{Task B Duluth : Logistic Regression, accuracy = .556, majority classifier = .598}
\label{table:taskbresults}
\end{table}

\begin{table}[h]
\centering
\begin{tabular}{l|l}
BC162  &   can’t believe it went from “idc i’ll eat ur ass” to “i’m actually not  
sexual anymore”\\ & b**ch i’m gonna kill {\bf u} \\
BC962  &   {\bf She} just wanna f*ck me cuz I be making money \\
BC406  &   {\bf B**ches} who got they shit together ain't bout to argue or be in drama everyday. \\
& That's a BROKE B**CH HOBBY! \\
BC233  & @USER {\bf Facebook} sucks ass, used to use it all the time under my real name up\\ 
& until 2017/2016 that's when it got unusable for me. \\
BC1152 & @USER {\bf It's} a place to shit and piss in \\
BC1253 OFF? &    being nice to people is not flirt {\bf u} dumb f*ck.\\
BC1233 OFF?  &  make me a playlist of songs {\bf you} wanna f*ck me to \\ 
BC1047 OFF? TIN?? &   @USER K increased my volume to hear this shit and I WAS NOT ALONE \\
BC1406 OFF?? TIN?? &   @USER The stupidity is immense, I'm sure their grandads 
will be spinning \\& around in tha graves.\\
BC963 OFF?? TIN?? &   @USER It's an ugly weapon, but war is ugly business.\\
\end{tabular}
\caption{Task B TIN-UNT category examples - gold standard indicates targeted insult while Duluth system predicts untargeted.}
\label{table:tinunt}
\end{table}

Given the significant level of disagreement between the gold standard and Duluth predictions we again carried out a qualitative analysis of misclassifications. We focused  on those cases where the gold standard said an offensive tweet was a targeted insult, and where the Duluth system classified it as untargeted. We refer to this as TIN-UNT. 

Table~\ref{table:tinunt} shows ten randomly selected examples from Task B. The first five shown seem to be correct as offensive targeted insults. However, we feel that there are questions about the other five. We do not find them obviously offensive, and we are uncertain that there is really anything being targeted in the last three instances (TIN??). We observed a similar distribution of question in a total of 10 random trials of 10 instances each. 

\subsection{Task C}


For Task C, the input is a targeted insult and the task is to decide if the target is an individual (IND), group (GRP) or other (OTHER). As can be seen in Table~\ref{table:taskcresults} the Duluth classifier's precision was relatively high for the IND category, and overall the classifier exceeded the majority baseline. 

\begin{table}[h]
\centering
\begin{tabular}{lrrrr|rrr}
     &  IND &  GRP &  OTH & & P & R & F1 \\ \hline
 IND &   537  &  69  &  47    & 653  & .822 & .926 & .871 \\
 GRP &   36  &  113  &  21    & 170 & .665 & .595 & .628\\
 OTH &    7 &  8   &    12  & 27 & .444 & .150 & .224 \\ \hline
     &   580  & 190   & 80    & 850 &  &  &  \\
\end{tabular}
\caption{Task C Duluth : Logistic Regression, accuracy = .779, majority classifier = .682}
\label{table:taskcresults}
\end{table}

In this task the most common confusion between the gold standard and Duluth predictions is that of tweets where the gold standard considers them as targeting a group (GRP) but the system predicts individual (IND). As we did for previous tasks we took several random trials (in this case only three given the smaller size of this class) and we show one of those trials in Table~\ref{table:grpind}.

\begin{table}[h]
\centering
\begin{tabular}{l|l}
BC518  & that whole ‘hot girl’ shit is just an excuse for b**ches to be they true selves \\
BC668  & @USER i didnt see that what the f*ck  ig stans are a disgusting breed \\
BC836 GRP?? & @USER @USER @USER Welp, I'm done here. F*ck off nazi. \\
BC1714 GRP??  &  @USER 238 followers tells me you’re the dumb F-u k making up Shiit \\
BC1803 OFF?     & Muted this idiot and his bandwagon of fools , awon oloshi alayi ni ironu \\
BC1347 OFF??    & Skinny dudes the worst to cuddle with.  Might as well fall asleep on a pile of remotes. \\
BC198  OFF?? & I like my men slightly clingy. Like be ambitious and motivated but once you’re \\
& off work be on my ass. \\
BC374 OFF??    & the f*ck is a bootycall i hate you all \\
BC1490? OFF?? & I hate a lazy b**ch , and b**ch isnt gender specific \\
BC1739 OFF?? & You can't make a n**ga loyal that's just some shit that can't be taught \\
BC659 OFF?? & i was at 422 followers and b**ches  thought it’d be cool to unfollow me  
lmao \\ 
& what the f*ck okay \\
\end{tabular}
\caption{Task C GRP-IND category examples - gold standard indicates targeted insult against group while Duluth system predicts against individual.}
\label{table:grpind}
\end{table}

Here we have significant questions about the majority of the tweets in the GRP-IND class. Only the first two instances appear to clearly be targeted insults against a group. Thereafter we have two instances where an individual is the target, and then six examples where is is unclear if the tweet is even offensive in the first place. It appears that much of the confusion may result over the use of the N-word and the B-word, as has been discussed for previous tasks. We suspect that the use of these terms may automatically make them offensive and targeted against a group. However, this is perhaps too coarse of a view and overlooks a great deal of nuance.

\section{Discussion}

The qualitative analysis described above suggests that any use of profanity often resulted in a tweet being labeled as offensive. This had a negative effect on the downstream tasks which took such tweets and tried to determine if the offense was targeted, and who was the target. 

We wonder if the use of .5 as the boundary between offensive or not, targeted or not, etc. may have been too lenient. By contrast, the Duluth system used a cutoff of .8. Our manual inspection suggested that tweets that scored above .8 tended to be somewhat harsher and more offensive than those with lower scores, although this is more of an intuition at this point rather than a conclusive finding. 


Table ~\ref{taskatraindist} and ~\ref{taskctraindist} (both in the Appendix) show the distribution of scores in the training data for Tasks A, B and C. These tables show the significance of the choice of .8 versus .5 as the category cutoff in that the total number of tweets that were considered offensive or targeted was much smaller with the .8 cutoff. Whether the tweets further down the tail are consistently more offensive is an interesting question for future work. 

\section{Ethical Considerations}

Identifying offensive language is a problem without a clear definition. The challenge is that a particular statement may or may not be offensive depending on the context in which it is used, and that depends on the nature of the source and target of such a statement. To further complicate matters, offensive statements may also be true statements. A totalitarian leader may be deeply offended by statements detailing human rights abuses committed by their regime, or a celebrity may be offended that a recent criminal charge is widely reported. 
There are also legitimately unclear boundaries. There are people who may be genuinely offended for religious or cultural reasons by any use of profanity. Since their preferences are genuine, should that be the standard that offensive language detection relies on? If so, the problem reduces to the simpler task of identifying profanity, which would then unfairly classify common everyday even friendly uses of profanity as offensive and subject it to flagging, removal, or other sanctions. 

It seems clear that we can't make blanket assumptions about what is offensive. Instead, we need to be very specific as to what our boundaries are for a particular corpus or task. In particular, we should be mindful of who is the target of such a language (as was done in both \cite{Offenseval2019} and \cite{Offenseval2020}), but we should also consider the source, and the context in which the language occurs. On a broader scale, we should reflect on who holds power \cite{Barbas2020}, and who gains and loses power if a statement is flagged as offensive.

While evaluation scores are important to advancing progress in NLP, there are some serious problems with making decisions about how well a problem is being solved simply by looking at such measures. We believe that in--depth qualitative analyses of mismatches between gold standard data and system predictions must be carried out in addition to providing more quantitative results. 

State of the art methods for many NLP problems have been shown to reach high levels of accuracy simply by learning spurious patterns in the training data without making a dent in the underlying problem we'd like to solve 
(e.g., \cite{niven-kao-2019-probing}).
We must acknowledge the possibility of this occurring in offensive language and hate speech detection and carry out qualitative analyses of our results in addition to the more typical quantitative ones. 

The danger of simply focusing on high accuracy is that we can be lulled into a false sense of success. In OffensEval--2020, 59 teams reached a macro F-1 score of .90 or better. Does this mean the problem of offensive language detection has been solved? We do not believe this is the case, nor do we think that any of the task participants would draw this conclusion. However, an outsider viewing such results might wrongly conclude that these methods and models are ready for use in the wild. This could result in a flawed  offensive language detector being deployed which would over or under identify problematic language, most likely to the disadvantage of already marginalized populations. 

\section{Conclusion}

This is the second year a Duluth system participated in OffensEval. Last year in OffensEval--2019 we took a relatively simple Machine Learning and rule-based approach and carried out an analysis of the results from the task. Our findings \cite{Pedersen19} were similar to this year, where the Duluth system did not rank particularly high. Also, we observed both in 2019 and 2020 that there appear to be some limitations in the gold standard annotations. In particular, in both years it seemed like there were quite a few false positives in the gold data, where tweets were labeled as offensive in Task A and potentially targeting in Task B when in fact they were not. It is not surprising that the 2019 OLID data and 2020 SOLID data would show similar characteristics, since OLID was the seed used for the distant supervision that created SOLID. 

In general it would appear virtually impossible to reliably annotate data without some background knowledge about the participants in the dialogue, as well as larger cultural contexts that might be at work (e.g, \cite{Patton2019,Frey2020}). That said we believe that annotated corpora is an important resource for this problem, and we need to continue to refine our processes for the creation of the same. In the creation of our own corpora we are working to develop Data Statements \cite{bender-friedman-2018-data} and plan to incorporate domain experts in the annotation process. 

\bibliography{tdp}
\bibliographystyle{coling}

\newpage
\appendix
\section{Appendix : Supplementary Tables}

\begin{table}[ht]
 \centering
 \begin{tabular}{c||rr|rr}
 \multicolumn{1}{c}{} & 
 \multicolumn{2}{c}{Task A} &
 \multicolumn{2}{c}{TaskB} \\ \hline
 range & percent & count & percent & count \\ \hline
 0 - .1 & 0.001 & 11,385 & 0.000 & 2\\
.1 - .2 & 0.284 & 2,580,150 & 0.103 & 19,479\\ \cline{4-5}
.2 - .3  & 0.351 & 3,189,564 & 0.409 & 77,312 \\
.3 - .4 & 0.135 & 1,229,725 &  0.155 & 29,212\\
.4 - .5 & 0.069 & 629,455 & 0.125 & 23,546\\
.5 - .6 & 0.044 & 404,462 & 0.114 & 21,599\\
.6 - .7 & 0.037 & 338,467 & 0.045 & 8,533\\
.7 - .8 & 0.038 & 349,121 & 0.043 & 8,174\\ \cline{2-3}
.8 - .9 & 0.039 & 351,169 & 0.006 & 1,118\\
.9 - 1.0 & 0.001 & 5,642 & 0.000 & 0 \\ \hline
 & 1.000 & 9.089,140 & 1.000 & 188,975\\
\end{tabular}
\caption{Training data score distribution Task A. Duluth system interpreted scores $\geq$ .8 as offensive (OFF), for Task B scores $\geq$ .2 indicated untargeted (UNT). Note that .5 was used as the cutoff for both tasks in the gold data.}
\label{taskatraindist}
\end{table}

\begin{table}[h]
 \centering
 \begin{tabular}{c||rr|rr|rr}
 \multicolumn{1}{c}{} &
 \multicolumn{2}{c}{IND} &
 \multicolumn{2}{c}{GRP} &
 \multicolumn{2}{c}{OTH}\\ \hline
 range & percent & count & percent & count & percent & count\\ \hline
0 - .1 & 0.004 & 703 & 0.027 & 5,143 & 0.018 & 3,382 \\
.1 - .2 & 0.042 & 7,871 & 0.462 & 87,276 & 0.342 & 64,700\\
.2 - .3 & 0.059 & 11,205 & 0.257 & 48,498 & 0.339 & 64,051\\
.3 - .4 & 0.082 & 15,563 &  0.114 & 21,581 & 0.223 & 42,171\\
.4 - .5 & 0.141 & 26,643 & 0.067 & 12,650 & 0.061 & 11,599 \\
.5 - .6 & 0.186 & 35,064 & 0.037 & 6,971 & 0.013 & 2,499\\
.6 - .7 & 0.195 & 36,942 & 0.023 & 4,345 & 0.003 & 513\\
.7 - .8 & 0.249 & 47,113 & 0.012 & 2,300 & 0.000 & 55\\ \cline{2-7}
.8 - .9 & 0.042 & 7,848 & 0.001 & 210 &  0.000 & 4\\
.9 - 1.0 & 0.000 & 22& 0.000 & 0 & 0.000 & 0 \\ \hline
 & 1.000 & 188,974 & 1.000 & 188,974 & 1.000 & 188,974\\
\end{tabular}
\caption{Training data score distribution Task C, higher scores indicate targeted individual (IND), group (GRP), or other (OTH). Duluth system interpreted scores $\geq$ .8 as IND, GRP, or OTH whereas gold data used .5 as the cutoff.}
\label{taskctraindist}
\end{table}


\begin{table}[h]
\centering
\begin{tabular}{l|l}
A3321 & are an ugly i hate straight your hetero frick fake \\
A3410  & Literally everyone wants a piece of my girlfriends ass. 
I need to treat her better smh. \\
A2257  & @USER Get the gains or die alone b**ch boy.\\
A2793  & @USER What a liar. White terrorists are his followers. \\
A406  & @USER f*ck off Deb. The planet is dying and you don't care \\
A171  & WAHT THE F**CK THEY ARE COMING TO LA f*ck THIS SHUT \\
& IM GOING TO WORK\\
A945 OFF?? & imagine thinking the death of a 20 year old man with an unborn child is funny ... \\
&  some people really are sick and pathetic\\
A217  OFF?? & @USER You mean because you're a coward \\
A3285  OFF?? & I wanna do the hot wing challenge shit look so good \\
A3456  OFF?? & Guys who start off the conversation sexually, you suck that’s just a turn off bye
\end{tabular}
\caption{Task A OFF-OFF category example : gold standard and Duluth system prediction agree on OFFensive. Those indicated with OFF?? may in fact be NOT.}
\label{table:offoff}
\end{table}

\begin{table}[h]
\centering
\begin{tabular}{l|l}
A2402 & @USER Idk, maybe they suddenly found out what a \\ 
& great person you are and decided to stan \\
A1113 & @USER it's not about chara, bby \\
A1271 & I remember when RBG broke her ribs MOST of what I saw \\ 
& from both sides of the isle were well wishes and hopes she would feel better soon. \\
A562 & @USER It was great \\
A1890 & @USER what’s up and thanks for the follow @USER says hello :) \\
A1813 & @USER @USER That would have made your other life easier. \\
A3501 & Great day to bless up \\
A1084 & Rest easy, Toni Morrison. You were the best of us. \\
A2228 & Click and Learn Some Valuable information! \\
A1738 & @USER you know the ones who aren't appreciated bro  it's okay \\
\end{tabular}
\caption{Task A NOT-NOT category examples : gold standard and Duluth system prediction agree on NOT offensive. Our analysis supports that these are NOT.}
\label{table:notnot}
\end{table}

\begin{table}[h]
\centering
\begin{tabular}{l|l}
A2193 & @USER @USER U call the whole Pastor a liar? \\
A1437 & @USER @USER @USER That sucks! \\
A996 & @USER UGH!  That sucks \\
A427 & Stop dissecting these sick manifestos, they do not  
point to anything but a deranged, \\
& mentally ill mind.\\
A1230 & my friend thinks my dad is a liar when really I'm a 
liar and he just helped me lie \\
A3818 & @USER @USER @USER The devil is a liar \\
A1884 & Wet pants. What the f*ck \\
A1082 & Fat, ugly and sick of it \\
A2060 & *at the zoo* these fursonas suck \\
A1480 NOT? & conspiracy theory : billie eilish farted on live for 
y’all weird ass grown men\\
& to stop sexualizing her
\end{tabular}
\caption{Task A NOT-OFF category examples - gold standard indicates NOT offensive while Duluth system predicts OFFensive. Our analysis suggests gold standard is correct except possibly for case indicated by NOT?.}
\label{table:notoff}
\end{table}
\end{document}